\documentclass[11pt]{article}
\usepackage{algorithm,algorithmic}
\usepackage{wrapfig} % for wrapping text around figures
\usepackage{sidecap}  % For putting caption beside figure
\usepackage{pgf,tikz,pgfplots}
\usetikzlibrary{arrows,automata,fit,positioning}
\usepackage{amssymb} % amssymb package, useful for mathematical symbols
\usepackage{amsfonts}
\usepackage{bm,amsbsy} % for bold symbols
\usepackage{amsmath}
\usepackage{stmaryrd}
\usepackage[normalem]{ulem}
\useunder{\uline}{\ul}{}
\usepackage{booktabs}
\usepackage{multirow}
\usepackage{tabularx}
\usepackage{caption}

\newcommand{\algoref}[1]{Algorithm~\ref{algo:#1}}  % use for citing algorithms

\newcommand{\Dat}{\mathcal{D}}

\newcommand{\Nrm}{\mathcal{N}}
\newcommand{\trp}{{^\top}} % transpose

\newcommand{\vphi}{\mathbf{\ensuremath{\bm{\phi}}}}

\newcommand{\vone}{\mathbf{1}} % one vector
\newcommand{\vz}{\mathbf{z}}
\newcommand{\vw}{\mathbf{w}}
\newcommand{\vn}{\mathbf{n}}
 
\newcommand{\vl}{\mathbf{\ensuremath{\bm{\mathit{l}}}}}
\newcommand{\vm}{\mathbf{\ensuremath{\bm{\mathit{m}}}}} 
\newcommand{\vs}{\mathbf{s}}

\newcommand{\vbeta}{\mathbf{\ensuremath{\bm{\beta}}}}

\newcommand{\vgamma}{\mathbf{\ensuremath{\bm{\gamma}}}}
\newcommand{\vlambda}{\mathbf{\ensuremath{\bm{\lambda}}}}
\newcommand{\vtheta}{\mathbf{\ensuremath{\bm{\theta}}}}
\newcommand{\vnu}{\mathbf{\ensuremath{\bm{\nu}}}}

\def\bbE{\mathbb{E}}

\newcommand{\mc}[1]{\mathcal{#1}}
\newcommand{\mb}[1]{\mathbf{#1}}

\newcommand{\appropto}{\mathrel{\vcenter{
  \offinterlineskip\halign{\hfil$##$\cr
    \propto\cr\noalign{\kern2pt}\sim\cr\noalign{\kern-2pt}}}}}

\renewcommand{\eqref}[1]{Eq.~\ref{eq:#1}}

\begin{document}

%------------------------------------------------------ title --------------------

\title{\textbf{Private Topic Modeling}}
\author{Mijung Park, {\it{Max Planck Institute for Intelligent Systems \& University of T\"ubingen}} \\
James Foulds, {\it{University of Maryland, Baltimore County}}\\
Kamalika Chaudhuri, {\it{University of California, San Diego}}\\
Max Welling, {\it{University of Amsterdam}}
}
 \date{}

\maketitle

\vspace{-0.5cm}

%------------------------------------------------------ abstract --------------------
\centerline{\textbf{Abstract}}
%\setspacing{1}
\hbox to \textwidth{\hrulefill} 

We develop a privatised stochastic variational inference method for Latent Dirichlet Allocation (LDA). The iterative nature of stochastic variational inference presents challenges: multiple iterations are required to obtain accurate posterior distributions, yet each iteration increases the amount of noise that must be added to achieve a reasonable degree of privacy. We propose a practical algorithm that overcomes this challenge by combining: (1) an improved composition method for differential privacy, called the {\it{moments accountant}}, which provides a tight bound on the privacy cost of multiple variational inference iterations and thus significantly decreases the amount of additive noise; and (2) privacy amplification resulting from subsampling of large-scale data. Focusing on conjugate exponential family models, in our private variational inference, all the posterior distributions will be privatised by simply perturbing expected sufficient statistics. Using Wikipedia data, we illustrate the effectiveness of our algorithm for large-scale data.

\hbox to \textwidth{\hrulefill}

\vspace{-0.1cm}

%Describe the problem or situation being addressed that led to your solution.
\section{Background}

\paragraph{Differential Privacy} (DP) is a formal
definition of the privacy properties of data analysis algorithms
\cite{Dwork14,dwork2006calibrating}. 
A randomized algorithm $\mathcal{M}(\mathbf{X})$ is said to
be $(\epsilon,\delta)$-differentially private if
\begin{equation}
Pr(\mathcal{M}(\mathbf{X}) \in \mathcal{S}) \leq \exp(\epsilon) Pr(\mathcal{M}(\mathbf{X}') \in \mathcal{S}) + \delta
\end{equation}
for all measurable subsets $\mathcal{S}$ of the range of $\mathcal{M}$ and for
all datasets $\mathbf{X}$, $\mathbf{X}'$ differing by a single entry (either by
excluding that entry or replacing it with a new entry).
 Here, an entry usually corresponds to a single individual's private value. If
$\delta = 0$, the algorithm is said to be $\epsilon$-differentially private,
and if $\delta > 0$, it is said to be {\it{approximately}} differentially
private. 
Intuitively, the definition states that the probability of any event does not
change very much when a single individual's data is modified, thereby limiting
the amount of information that the algorithm reveals about any one individual.
We observe that $\mathcal{M}$ is a randomized algorithm, and randomization is
achieved by either adding external noise, or by subsampling.  In this paper, we use the ``\emph{include/exclude}'' version of DP, in which {\it{differing by a single entry}} refers to the inclusion or exclusion of that entry in the dataset.

\paragraph{Variational inference for the conjugate exponential models}
Variational inference is an optimization-based posterior inference method, which simplifies to a two-step procedure when the model falls into the Conjugate-Exponential (CE) class of models. CE family models satisfy two conditions \cite{Beal_03}:
\begin{align}
&(1)\mbox{ The complete-data likelihood is in the exponential family}: \nonumber \\
& \qquad p(\Dat_n, \vl_n| \vm) = g(\vm) f(\Dat_n, \vl_n) \exp(\vn(\vm)\trp \vs(\Dat_n, \vl_n)), \\
&(2)\mbox{ The prior over $\vm$ is conjugate to the complete-data likelihood}: \nonumber \\
& \qquad p(\vm|\tau, \vnu) = h(\tau, \vnu) g(\vm)^{\tau} \exp(\vnu\trp\vn(\vm)).
\end{align} where natural parameters and sufficient statistics of the complete-data likelihood are denoted by  $\vn(\vm)$ and $\vs(\Dat_n, \vl_n)$, respectively, and $g,f,h$ are some known functions. The hyperparameters are denoted by $\tau$ (a scalar) and $\vnu$ (a vector).

The  variational inference algorithm for a CE family model optimises the lower bound on the model log marginal likelihood given by,
\begin{equation}\label{eq:var_lbd}
\mathcal{L}(q(\vl) q(\vm)) = \int d\vm \; d\vl \; q(\vl) q(\vm) \log \frac{p(\vl, \Dat, \vm)}{q(\vl) q(\vm)}, %JF: I changed \mathcal{L}(q(\vl), q(\vm)) to \mathcal{L}(q(\vl) q(\vm)) to be consistent with my earlier notation.
\end{equation}
where we assume that the joint approximate posterior distribution over the latent variables and model parameters $q(\vl, \vm)$ is factorised via the mean-field assumption as 
$
q(\vl, \vm) = q(\vl) q(\vm) = q(\vm) \prod_{n=1}^N q(\vl_n),
$
and that each of the variational distributions  also has the form of an exponential family distribution.
Computing the derivatives of the variational lower bound in  \eqref{var_lbd} with respect to each of these variational distributions and setting them to zero yield the following two-step procedure. 
%\vspace{-0.05cm}
{\small\begin{align}\label{eq:VB}
&(1) \mbox{ First, given expected natural parameters $\bar{\vn}$, the E-step computes:}    \nonumber \\
& \qquad  q(\vl) = \prod_{n=1}^N q(\vl_n) \propto  \prod_{n=1}^N f(\Dat_n, \vl_n) \exp(\bar{\vn}\trp \vs(\Dat_n, \vl_n)) = \prod_{n=1}^N p(\vl_n|\Dat_n, \bar{\vn}) .  \\
& \qquad \mbox{Using $q(\vl)$, it outputs expected sufficient statistics,  the  expectation  of  ${\vs}(\Dat_n, \vl_n)$}\nonumber \\
& \qquad \mbox{with probability density $q(\vl_n)$ : } \bar{\vs}(\Dat) = \tfrac{1}{N} \sum_{n=1}^N \langle {\vs}(\Dat_n, \vl_n) \rangle_{q(\vl_n)}. \nonumber \\
&(2)\mbox{ Second, given expected sufficient statistics $\bar{\vs}(\Dat)$, the M-step computes:}  \nonumber \\
&  \qquad q(\vm) =  h(\tilde{\tau}, \tilde{\vnu}) g(\vm)^{\tilde{\tau}} \exp(\tilde{\vnu} \trp \vn(\vm)), \mbox{ where } \tilde{\tau} = \tau + N, \; \tilde{\vnu} = \vnu + N \bar{\vs}(\Dat).\\
& \qquad \mbox{Using $q(\vm)$, it outputs expected natural parameters $\bar{\vn}= \langle \vn(\vm) \rangle_{q(\vm)}$}.\nonumber
\end{align} \normalsize}

\section{Privacy preserving VI algorithm for CE family}
The only place where the algorithm looks at the data is when computing the expected sufficient statistics $\bar{\vs}(\Dat)$ in the first step. The expected sufficient statistics then dictates the expected natural parameters in the second step. So, perturbing the sufficient statistics leads to perturbing both posterior distributions $q(\vl)$ and $q(\vm)$. Perturbing sufficient statistics in exponential families is also used in \cite{DPEM16}. Existing work focuses on privatising posterior distributions in the context of posterior sampling \cite{zhang2016differential,dimitrakakis2014robust,FGWC16,BartheFGAGHS16}, while our work focuses on privatising approximate posterior distributions for optimisation-based approximate Bayesian inference.
Suppose there are two neighbouring datasets $\Dat$ and $\Dat'$, where there is only one datapoint difference among
them. We also assume that the dataset is pre-processed such that the L2 norm of any datapoint is less
than 1. The maximum difference in the expected sufficient statistics given the datasets, e.g., the L-1
sensitivity of the expected sufficient statistics is given by (assuming s is a vector of length L) 
$\Delta\vs = \max_{\Dat, \Dat', q(\vl), q(\vl')} \sum_{l=1}^L\frac{1}{N}|\mathbb{E}_{q(\vl)}\vs_l(\Dat,\vl)-\mathbb{E}_{q(\vl')}\vs_l(\Dat',\vl')|$.Under some models like LDA below, 
expected sufficient statistic has a limited sensitivity, in which case we add noise to each coordinate of the expected sufficient statistics to compensate the maximum change.

\section{Privacy preserving latent Dirichlet allocation (LDA)}
The most successful topic modeling is based on LDA, where the generative process is given by \cite{blei2003latent}.
Its generative process is given by
%\vspace{-0.2cm}
%\small{
\begin{itemize}
\item Draw topics $\vbeta_k \sim $ Dirichlet $(\eta \vone_{V})$, for $k=\{1,\ldots, K\}$, where $\eta$ is a scalar hyperarameter.
\item For each document $d \in \{ 1, \ldots, D \}$
\begin{itemize}
\item Draw topic proportions $ \vtheta_d \sim$ Dirichlet $(\alpha \vone_{K})$, where $\alpha$ is a scalar hyperarameter.
\item For each word $n \in \{ 1, \ldots, N \}$
\begin{itemize}
\item Draw topic assignments $\vz_{dn} \sim$ Discrete$(\vtheta_d)$
\item Draw word $\vw_{dn} \sim$ Discrete$(\vbeta_{\vz_{dn}})$
\end{itemize}
\end{itemize}
\end{itemize} 
%}\normalsize
%\vspace{-0.2cm}
where each observed word is represented by an indicator vector $\vw_{dn}$ ($n$th word in the $d$th document) of length $V$, and where $V$ is the number of terms in a fixed vocabulary set. The topic assignment latent variable $\vz_{dn}$ is also an indicator vector of length $K$, where $K$ is the number of topics. 
The LDA model falls in the CE family, viewing $\vz_{d, 1:N}$ and  $\vtheta_d $ as two types of latent variables: $\vl_d = \{ \vz_{d, 1:N}, \vtheta_d \}$, and $\vbeta$ as model parameters $\vm = \vbeta$. The conditions for CE are satisfied: (1) the complete-data likelihood is in exponential family: 
$p(\vw_{d, 1:N}, \vz_{d, 1:N}, \vtheta_d| \vbeta) \propto  f(\Dat_d, \vz_{d, 1:N}, \vtheta_d) \exp ( \sum_{n}\sum_{k} [\log \vbeta_k] \trp [\vz_{dn}^k \vw_{dn} ] ),$
where   $f(\Dat_d, \vz_{d, 1:N}, \vtheta_d) \propto\exp([\alpha \vone_K] \trp [\log \vtheta_d] + \sum_{n}\sum_{k}\vz_{dn}^k \log \vtheta_{d}^k) $; and
(2) we have a conjugate prior over $\vbeta_k$:  
$p(\vbeta_k|\eta \vone_{V}) \propto \exp([\eta \vone_{V}]\trp [\log \vbeta_k]), $
 for  $k=\{1, \ldots, K\}$.
%\end{itemize}
%\end{align} 
%\vspace{-0.2cm}
For simplicity, we assume hyperparameters $\alpha$ and $\eta$ are set manually. 
Under the LDA model, we assume the variational posteriors are given by 
\begin{itemize}
\item Discrete : $q(\vz_{dn}^k|\vphi_{dn}^k) \propto  \exp(\vz_{dn}^k \log \vphi_{dn}^k)$, with variational parameters for capturing the posterior topic assignment, 
$
\vphi_{dn}^k \propto \exp(\langle \log \vbeta_k \rangle_{q(\vbeta_k)} \trp \vw_{dn} + \langle \log \vtheta_d^k \rangle_{q(\vtheta_d)}).
$
\item Dirichlet : $q(\vtheta_d | \vgamma_d) \propto \exp( \vgamma_d \trp \log \vtheta_d), \mbox{where  } \vgamma_d =  \alpha \vone_K + \sum_{n=1}^N \langle \vz_{dn} \rangle_{q(\vz_{dn})}$, 
\end{itemize} where these two distributions are computed in the E-step behind the privacy wall.  
 The expected sufficient statistics are $\bar{\vs}_k^v = \tfrac{1}{D}\sum_{d} \sum_{n} \langle \vz_{dn}^k \rangle_{q(\vz_{dn})} \vw^v_{dn} = \tfrac{1}{D}\sum_{d} \sum_{n} \vphi_{dn}^k \vw_{dn}$.
Then, in the M-step, we compute the posterior 
\begin{itemize}
\item Dirichlet : $q(\vbeta_k|\vlambda_k  ) \propto \exp( \vlambda_k \trp \log \vbeta_k), 
 \mbox{ where }  \vlambda_k = \eta \vone_V + \sum_{d} \sum_{n} \langle \vz_{dn}^k \rangle_{q(\vz_{dn})} \vw_{dn} $.
\end{itemize}

\paragraph{Sensitivity analysis}
In a large-scale data setting, it is impossible to handle the entire dataset at once. In such case, stochastic learning using noisy sufficient statistics computed on mini-batches of data. At each learning step with a freshly drawn mini-batch of data (size $S$), we perturb the expected sufficient statistics. While each document has a different document length $N_d$, we limit the maximum length of any document to $N$ by randomly selecting $N$ words in a document if the number of words in the document is longer than $N$.

We add Gaussian noise to each component of the expected sufficient statistics, which is a matrix of size $K \times V$,
\begin{align}\label{eq:gaussian_LDA}
\tilde{\bar{\vs}}_{k}^v = {\bar{\vs}}_{k}^v + Y_k^v, \mbox{ where } Y_k^v \sim \Nrm(0, \sigma^2 (\Delta \bar{\vs})^2), 
\end{align}
where %$\vs_{k}^v$ is the $v$th coordinate of a vector of length $V$, %JF: I don't know why we need this statement, and actually it seems that we are treating the whole sufficient statistic matrix as a vector for the purposes of sensitivity.  So I deleted this.
$\bar{\vs}_{k}^v = \tfrac{1}{S} \sum_d \sum_n \vphi_{dn}^k \vw_{dn}^v$, and $\Delta \bar{\vs}$ is the sensitivity.  We then map the perturbed components to 0 if they become negative.  For LDA, the worst-case sensitivity is given by
{\small{\begin{align}\label{eq:sen_lap_lda}
\Delta \bar{\vs} 
&= \max_{|\Dat \setminus {\Dat'}|=1} \sqrt{ \sum_k \sum_v (\bar{\vs}_{k}^v(\Dat)-\bar{\vs}_{k}^v(\Dat'))^2}, \nonumber \\
&= \max_{|\Dat \setminus {\Dat'}|=1}  \sqrt{\sum_k \sum_v  \left( \frac{1}{
S}\sum_n \sum_{d=1}^S \vphi_{dn}^k \vw_{dn}^v - \frac{1}{S-1} \sum_n \sum_{d=1}^{S-1}\vphi_{dn}^k \vw_{dn}^v \right)^2}, \nonumber \\
 &=  \max_{|\Dat \setminus {\Dat'}|=1}  \sqrt{\sum_k \sum_v \left|\frac{S-1}{S} \frac{1}{S-1} \sum_n \sum_{d=1}^{S-1} \vphi_{dn}^k \vw_{dn}^v + \frac{1}{S} \sum_n \vphi_{Sn}^k \vw_{Sn}^v - \frac{1}{S-1} \sum_n \sum_{d=1}^{S-1}\vphi_{dn}^k \vw_{dn}^v \right|^2}, \nonumber \\
 %\mbox{ due to \eqref{gen_sen}} \nonumber \\
  &=  \max_{\vphi_{Sn}^k, \vw_{Sn}^v}  \sqrt{\sum_k \sum_v \left|\frac{1}{S} \sum_n \vphi_{Sn}^k \vw_{Sn}^v -\frac{1}{S} \left( \frac{1}{S-1} \sum_n \sum_{d=1}^{S-1}\vphi_{dn}^k \vw_{dn}^v\right) \right|^2}, \nonumber \\
    &=  \max_{\vphi_{Sn}^k, \vw_{Sn}^v}  \sqrt{\sum_k \sum_v \left|\frac{1}{S} \sum_n \vphi_{Sn}^k \vw_{Sn}^v\right|^2}, \nonumber \\
    & \quad \mbox{ since $0 \leq \vphi_{dn}^k \leq 1$, $\vw_{dn}^v \in \{0, 1 \}$, and we assume $0 \leq \frac{1}{S-1} \sum_n \sum_{d=1}^{S-1}\vphi_{dn}^k \vw_{dn}^v \leq \sum_n \vphi_{Sn}^k \vw_{Sn}^v $,  }\nonumber \\
 &\leq \max_{\vphi_{Sn}^k, \vw_{Sn}^v} \; \; \frac{1}{S} \sum_n (\sum_k\vphi_{Sn}^k) (\sum_v \vw_{Sn}^v)  \leq \frac{N}{S}, \end{align}}}\normalsize
since  $\sum_k\vphi_{Sn}^k = 1$, and $\sum_v \vw_{Sn}^v=1$.  This sensitivity accounts for the worst case in which all $NS$ words in the minibatch are assigned to the same entry of $\bar{\vs}$, i.e. they all have the same word type $v$, and are hard-assigned to the same topic $k$ in the variational distribution.  In our practical implementation, we improve the sensitivity by exploiting the fact that most typical sufficient statistic matrix $\bar{\vs}$ given a minibatch (where the size of the matrix is the number of topics by the number of words in the vocabulary set) has a much smaller norm than this worst case.  Specifically, inspired by \cite{2016arXiv160700133A}, we apply a norm clipping strategy, in which the matrix $\bar{\vs}$ is clipped (or projected) such that the Frobenious norm of the matrix is bounded by $|\bar{\vs}| \leq a \frac{N}{S}$, for a user-specified $a \in (0,1]$.  For each minibatch, if this criterion is not satisfied, we project the expected sufficient statistics down to the required norm via
\begin{equation}
\bar{\vs} := a \frac{N}{S}\frac{\bar{\vs}}{{|\bar{\vs}|}} \mbox{ .} 
\end{equation}
After this the procedure, the sensitivity of the entire matrix becomes $a \Delta \bar{\vs}$ (i.e., $a N/S$), and we add noise on this scale to the clipped expected sufficient statistics.  We set $a = 0.1$ in our experiments, which empirically resulted in clipping being applied to around $3/4$ of the documents.
The resulting algorithm is summarised in \algoref{PPVI_LDA_minibatches}.

\begin{algorithm}[t]
\caption{Private LDA}
\label{algo:PPVI_LDA_minibatches}
\begin{algorithmic}
\REQUIRE Data $\Dat$. Define $D$ (documents), $V$ (vocabulary), $K$ (number of topics).\\
\quad \quad \;   Define $\rho_{t} = (\tau_0 + t)^{-\kappa}$, mini-batch size $S$, hyperparameters $\alpha, \eta$, $\sigma^2$, and $a$
\ENSURE Privatised expected natural parameters $\langle \log \vbeta_k \rangle_{q(\vbeta_k)}$ and sufficient statistics $\tilde{\bar{\vs}}$.
%\STATE Compute the per-iteration privacy budget ($\epsilon_{iter}, \delta_{iter}$) using Table 1.
\STATE Compute the sensitivity of the expected sufficient statistics given in \eqref{sen_lap_lda}.
\FOR{$t = 1, \ldots, J $}
\STATE {\it{\textbf{(1) E-step}}}: Given expected natural parameters $\langle \log \vbeta_k \rangle_{q(\vbeta_k)}$
\FOR{$d=1,\ldots,S$}
   \STATE Compute $q(\vz_{dn}^k)$ parameterised by $\vphi_{dn}^k \propto \exp(\langle \log \vbeta_k \rangle_{q(\vbeta_k)} \trp \vw_{dn} + \langle \log \vtheta_d^k \rangle_{q(\vtheta_d)})$.
   \STATE Compute $q(\vtheta_d)$ parameterised by $\vgamma_d = \alpha \vone_K + \sum_{n=1}^N \langle \vz_{dn}\rangle_{q(\vz_{dn})}$.
\ENDFOR
\STATE Compute the expected sufficient statistics $\bar{\vs}_k^v = \tfrac{1}{S}\sum_d \sum_n \vphi_{dn}^k \vw_{dn}^v$.
\IF{$|\bar{\vs}| > aN/S$}
\STATE $\bar{\vs} := a \frac{ N}{S}\frac{\bar{\vs}}{{|\bar{\vs}|}}$ 
\ENDIF
\STATE Output the perturbed expected sufficient statistics $ \tilde{\bar{\vs}}_k^v = \bar{\vs} + Y_k^v$, where $Y_k^v$ is Gaussian noise given in \eqref{gaussian_LDA}, but using sensitivity $aN/S$.
%\STATE Output the perturbed expected sufficient statistics $ \tilde{\bar{\vs}}_k^v = \tfrac{1}{S}\sum_d \sum_n \vphi_{dn}^k \vw_{dn}^v + Y_k^v$, where $Y_k^v$ is Gaussian noise given in \eqref{gaussian_LDA}.
\STATE Clip negative entries of $\tilde{\bar{\vs}}$ to 0.
\STATE Update the log-moment functions.
\STATE {\it{\textbf{(2) M-step}}}: Given perturbed expected sufficient statistics $ \tilde{\bar{\vs}}_k $,
\STATE Compute $q(\vbeta_k)$ parameterised by $\vlambda_k^{(t)} = \eta \vone_V + D \tilde{\bar{\vs}}_k$.
\STATE Set $\vlambda^{(t)} \mapsfrom (1-\rho_t)\vlambda^{(t-1)} + \rho_t \vlambda^{(t)} $.
\STATE Output expected natural parameters $\langle \log \vbeta_k \rangle_{q(\vbeta_k)}$.
\ENDFOR
\end{algorithmic}
\end{algorithm}

\section{Privacy analysis of private LDA}

We use the {\it{Moments Accountant}} (MA) composition method 
\cite{2016arXiv160700133A} for accounting for privacy loss incurred by
successive iterations of an iterative mechanism. We choose this method as it provides
tight privacy bounds (cf., \cite{dwork2010boosting}).  
The moments accountant method is based on the concept of a {\it{privacy loss
random variable}}, which allows us to consider the entire spectrum of
likelihood ratios $\frac{Pr(\mathcal{M}(\mathbf{X}) =
o)}{Pr(\mathcal{M}(\mathbf{X}') = o)}$ induced by a privacy mechanism
$\mc{M}$. Specifically, the privacy loss random variable corresponding to a
mechanism $\mathcal{M}$, datasets $\mathbf{X}$ and $\mathbf{X'}$, and an
auxiliary parameter $w$ is a random variable defined as follows:
$L_{\mc{M}}(\mb{X}, \mb{X'}, w) := \log \frac{Pr(\mathcal{M}(\mathbf{X}, w) = o)}{Pr(\mathcal{M}(\mathbf{X}', w) = o)}, \quad {\text{with likelihood\ }} Pr(\mathcal{M}(\mathbf{X}, w) = o),$
where $o$ lies in the range of $\mc{M}$. Observe that if $\mc{M}$ is
$(\epsilon, 0)$-differentially private, then the absolute value of
$L_{\mc{M}}(\mb{X}, \mb{X'}, w)$ is at most $\epsilon$ with probability $1$. 

The moments accountant method exploits properties of this privacy loss random
variable to account for the privacy loss incurred by applying mechanisms
$\mc{M}_1, \ldots, \mc{M}_t$ successively to a dataset $\mb{X}$; this is done
by bounding properties of the log of the moment generating function of the
privacy loss random variable. Specifically, the log moment function
$\alpha_{\mc{M}_t}$ of a mechanism $\mc{M}_t$ is defined as:
\begin{equation}\label{eqn:logmomentfn}
\alpha_{\mc{M}_t}(\lambda) = \sup_{\mb{X}, \mb{X'}, w} \log \bbE[ \exp(\lambda  L_{\mc{M}_t}(\mb{X}, \mb{X'}, w))],
\end{equation}
where $\mb{X}$ and $\mb{X'}$ are datasets that differ in the private value of a
single person. \cite{2016arXiv160700133A} shows that if $\mc{M}$ is the combination of mechanisms $(\mc{M}_1, \ldots, \mc{M}_k)$ where each mechanism addes independent noise, then, its log moment generating function $\alpha_{\mc{M}}$ has the property that:
\begin{equation}\label{eqn:logmomentadditive}
\alpha_{\mc{M}}(\lambda) \leq \sum_{t=1}^{k} \alpha_{\mc{M}_t}(\lambda)
\end{equation}
Additionally, given a log moment function $\alpha_{\mc{M}}$, the corresponding mechanism $\mc{M}$ satisfies a range of privacy parameters $(\epsilon, \delta)$ connected by the following equation:
%and a target $\epsilon$ or $\delta$, the privacy parameters of the corresponding mechanism can be recovered by solving:
\begin{equation}\label{eqn:logmomenttoprivacy}
\delta = \min_{\lambda} \exp(\alpha_{\mc{M}}(\lambda) - \lambda \epsilon)
\end{equation}

These properties immediately suggest a procedure for tracking privacy loss incurred by a combination of mechanisms $\mc{M}_1, \ldots, \mc{M}_k$ on a dataset. For each mechanism $\mc{M}_t$, first compute the log moment function $\alpha_{\mc{M}_t}$; for simple mechanisms such as the Gaussian mechanism this can be done by simple algebra. Next, compute $\alpha_{\mc{M}}$ for the combination $\mc{M} = (\mc{M}_1, \ldots, \mc{M}_k)$ from~(\ref{eqn:logmomentadditive}), and finally, recover the privacy parameters of $\mc{M}$ using~(\ref{eqn:logmomenttoprivacy}) by either finding the best $\epsilon$ for a target $\delta$ or the best $\delta$ for a target $\epsilon$. In some special cases such as composition of $k$ Gaussian mechanisms, the log moment functions can be calculated in closed form; the more common case is when closed forms are not available, and then a grid search may be performed over $\lambda$. 

In Algorithm 1, observe that iteration $t$ of the algorithm subsamples a $\nu = S/D$ fraction of the dataset, computes the sufficient statistics based on this subsample, and perturbs it using the Gaussian mechanism with variance $\sigma^2 I_d$. To simplify the privacy calculations, we assume that each example in the dataset is included in a minibatch according to an independent coin flip with probability $\nu$.  From Proposition 1.6 in \cite{zCDP16} along with simple algebra, the log moment function of the Gaussian Mechanism $\mc{M}$ applied to a query with $L_2$-sensitivity $\Delta$ is $\alpha_{\mc{M}}(\lambda) = \frac{\lambda (\lambda + 1) \Delta^2}{2 \sigma^2}$. To compute the log moment function for the subsampled Gaussian Mechanism, we follow \cite{2016arXiv160700133A}. Let $\beta_0$ and $\beta_1$ be the densities $\mc{N}(0, (\sigma/\Delta)^2)$ and $\mc{N}(1, (\sigma/\Delta)^2)$, and let $\beta = (1 - \nu) \beta_0 + \nu \beta_1$ be the mixture density; then, the log moment function at $\lambda$ is $\max \log (E_1, E_2)$ where $E_1 = \bbE_{z \sim \beta_0} [ (\beta_0(z)/\beta(z))^{\lambda}]$ and $E_2 = \bbE_{z \sim \beta} [ (\beta(z)/\beta_0(z))^{\lambda}]$. $E_1$ and $E_2$ can be numerically calculated for any $\lambda$, and we maintain the log moments over a grid of $\lambda$ values. 

Note that our algorithms are run for a prespecified number of iterations, and with a prespecified $\sigma$; this ensures a certain level of ($\epsilon,\delta$) guarantee in the released expected sufficient statistics from Algorithm 1. 

\begin{figure}[t]
	\centering
	\centerline{\includegraphics[width=0.75\textwidth]{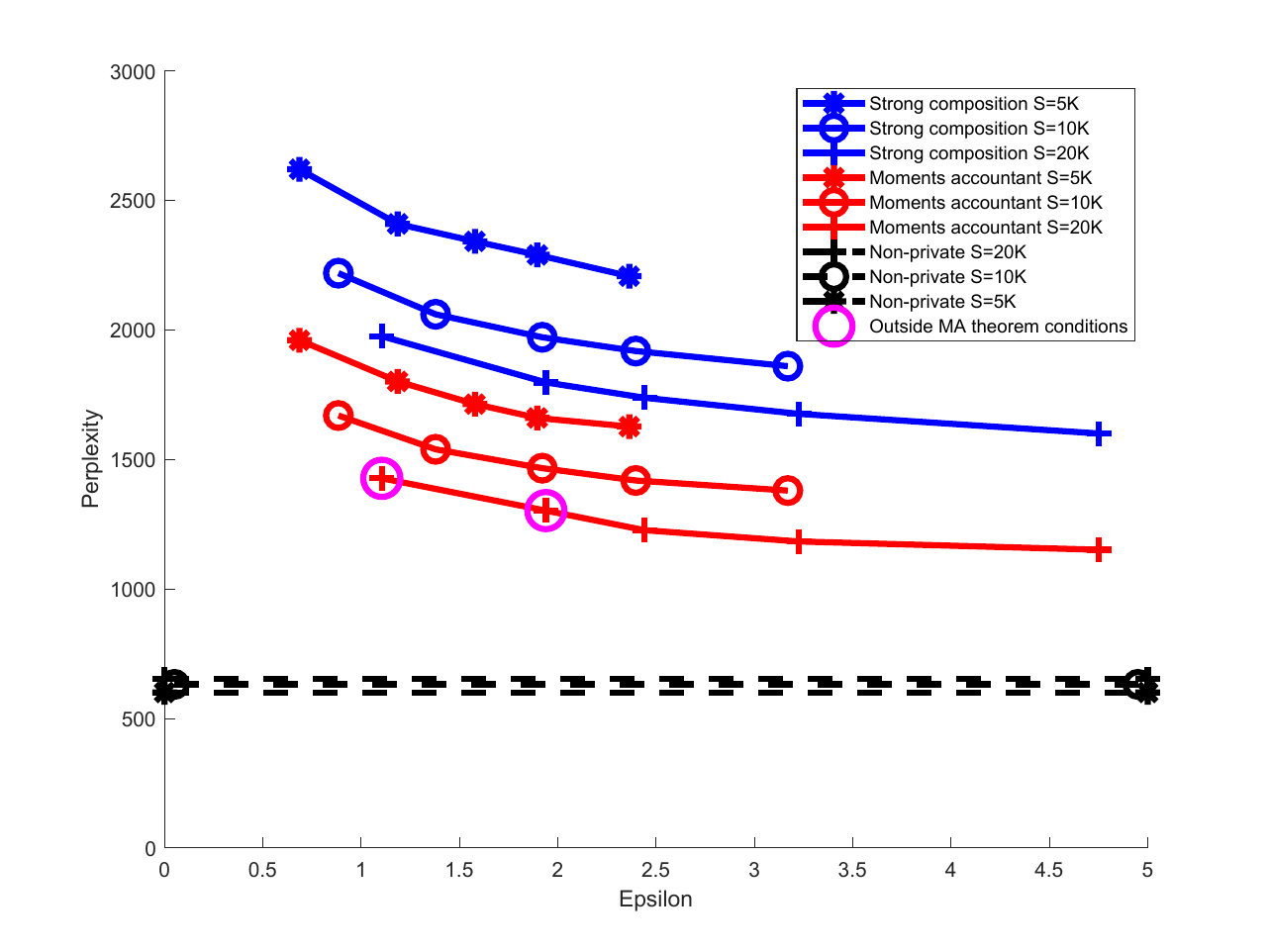}}
	\caption{ Epsilon versus perplexity, varying $\sigma$ and $S$, Wikipedia data, one epoch.  The parameters for the two data points indicated by the pink circles do not satisfy the conditions of the moments accountant composition theorem, so those $\epsilon$ values are not formally proved.	}
	\label{fig:epsilonVsPerplexity}
\end{figure}

\section{Experiments using Wikipedia data}

%We downloaded a random $D=400,000$ documents from Wikipedia.  We then tested our VIPS algorithm on the Wikipedia dataset with three different values of total privacy budget $\epsilon =\{1.2, 2.3, 4.6\}$ by varying values of $\sigma$, using different mini-batch sizes $S=\{10, 100, 200\}$, until the algorithm sees up to $160,000$ documents.  We assumed there are $100$ topics, and we used a vocabulary set of approximately $8000$ terms. 
%%Under this setting, we first plotted the per-iteration privacy budget as a function of the number of documents seen in \figref{budget_plot}, assuming that the total privacy budget is 1. Note that larger number of documents imply longer iterations. Due to privacy amplification, it is more beneficial to decrease the amount of noise to add when the mini-batch size is small. CDP composition results in the highest per-iteration privacy budget and Linear composition is the worst (lin). 
We downloaded a random $D=400,000$ documents from Wikipedia to test our algorithm.  We used $50$ topics and a vocabulary set of approximately $8000$ terms.  The algorithm was run for one epoch in each experiment.%JF: New LDA experiments

We compared our moments accountant approach with a baseline method using the {\it{strong composition}}
%First, in {\it{linear}} (Lin) composition (Theorem 3.16 of \cite{Dwork14}), privacy degrades linearly with the number of iterations. This result is from the Max Divergence of the privacy loss random variable being bounded by a total budget. Hence, the linear composition yields ($J\epsilon'$, $J\delta'$)-DP. We use \eqref{amplif_per_tot} to map ($\epsilon', \delta'$) to $(\epsilon_{iter}, \delta_{iter})$.
%
(Theorem 3.20 of \cite{Dwork14}), resulting from the max divergence of the privacy loss random variable being bounded by a total budget including a slack variable $\delta$, which yields
$(J\epsilon' (e^{\epsilon'}-1) + \sqrt{2J\log(1/\delta{''})}\epsilon', \; \delta{''} + J\delta')$-DP.

As our evaluation metric, we compute an upper bound on the perplexity on held-out documents.  Perplexity is an information-theoretic measure of the predictive performance of probabilistic models which is commonly used in the context of language modeling \cite{jelinek1977perplexity}.  The perplexity of a probabilistic model $p_{model}(x)$ on a test set of $N$ data points $x_i$ (e.g. words in a corpus) is defined as
\begin{equation}
\mbox{perplexity}(\Dat^{test}, \vlambda) \leq \exp \left[ - \left(\sum_i \langle \log p(\vn^{test}, \vtheta_i, \vz_{i}|\vlambda) \rangle_{q(\vtheta_i, \vz_i)} -  \langle \log q(\vtheta, \vz) \rangle_{q(\vtheta, \vz)} \right) / \sum_{i,n} \vn_{i,n}^{test} \right], \nonumber 
\end{equation}
 where $\vn_i^{test}$ is a vector of word counts for the $i$th document, $\vn^{test} = \{ \vn_i^{test} \}_{i=1}^I$. In the above, we use the $\vlambda$ that was calculated during training. We compute the posteriors over $\vz$ and $\vtheta$ by performing the first step in our algorithm using the test data and the perturbed sufficient statistics we obtain during training. We adapted the python implementation by the authors of \cite{NIPS2010_3902} for our experiments.%The per-word-perplexity is shown in \figref{conv}.
 
Figure \ref{fig:epsilonVsPerplexity} shows the trade-off between $\epsilon$ and per-word perplexity on the Wikipedia dataset for the different methods under a variety of conditions, in which we varied the value of $\sigma \in \{1.0, 1.1,1.24,1.5,2\}$ and the minibatch size $S \in \{5,000, 10,000, 20,000\}$.  We found that the moments accountant composition substantially outperformed strong composition in each of these settings.  Here, we used relatively large minibatches, which were necessary to control the signal-to-noise ratio in order to obtain reasonable results for private LDA. Larger minibatches thus had lower perplexity.  However, due to its impact on the subsampling rate, increasing $S$ comes at the cost of a higher $\epsilon$ for a fixed number of documents processed (in our case, one epoch).  The minibatch size $S$ is limited by the conditions of the moments accountant composition theorem shown by \cite{2016arXiv160700133A}, with the largest valid value being obtained at around $S\approx20,000$ for the small noise regime where $\sigma \approx 1$.  

In Table \ref{my-table}, for each method, we show the top $10$ words in terms of assigned probabilities for $3$ example topics. %Due to space limit, %Jimmy: we don't have a space limit in this version!
Non-private LDA results in the most coherent words among all the methods. %For the private LDA models with a total privacy budget $\epsilon=2.3$ ($S=100, J=1600$), as we move from moments accountant to strong composition, the amount of noise added gets larger, and therefore the topics become less coherent.
For the private LDA models with a total privacy budget $\epsilon=2.44$ ($S=20,000, \sigma=1.24$), as we move from moments accountant to strong composition, the amount of noise added gets larger, and the topics become less coherent.  We also observe that the probability mass assigned to the most probable words decreases with the noise, and thus strong composition gave less probability to the top words compared to the other methods.

\begin{table}[t]
	\centering
	\caption{Posterior topics from private ($\epsilon=2.44$) and non-private LDA}
	\label{my-table}
	\begin{tabular}{llllll}
		\toprule
		Non-private &        & Moments Acc. &        & Strong Comp. &        \\ %\midrule
		\cmidrule(r){1-2} \cmidrule(l){3-4} \cmidrule(l){5-6}
		topic 33:   &        & topic 33:   &        & topic 33:   &        \\
		german      & 0.0244 & function    & 0.0019 & resolution  & 0.0003 \\
		system      & 0.0160 & domain      & 0.0017 & northward   & 0.0003 \\
		group       & 0.0109 & german      & 0.0011 & deeply      & 0.0003 \\
		based       & 0.0089 & windows     & 0.0011 & messages    & 0.0003 \\
		science     & 0.0077 & software    & 0.0010 & research    & 0.0003 \\
		systems     & 0.0076 & band        & 0.0007 & dark        & 0.0003 \\
		computer    & 0.0072 & mir         & 0.0006 & river       & 0.0003 \\
		software    & 0.0071 & product     & 0.0006 & superstition& 0.0003 \\
		space       & 0.0061 & resolution  & 0.0006 & don         & 0.0003 \\
		power       & 0.0060 & identity    & 0.0005 & found       & 0.0003 \\
		&        &             &        &             &        \\
		topic 35:   &        & topic 35:   &        & topic 35:   &        \\
		station     & 0.0846 & station     & 0.0318 & station     & 0.0118 \\
		line        & 0.0508 & line        & 0.0195 & line        & 0.0063 \\
		railway     & 0.0393 & railway     & 0.0149 & railway     & 0.0055 \\
		opened      & 0.0230 & opened      & 0.0074 & opened      & 0.0022 \\
		services    & 0.0187 & services    & 0.0064 & services    & 0.0015 \\
		located     & 0.0163 & closed      & 0.0056 & stations    & 0.0015 \\
		closed      & 0.0159 & code        & 0.0054 & closed      & 0.0014 \\
		owned       & 0.0158 & country     & 0.0052 & section     & 0.0013 \\
		stations    & 0.0122 & located     & 0.0051 & platform    & 0.0012 \\
		platform    & 0.0109 & stations    & 0.0051 & company     & 0.0010 \\
%		&        &             &        &             &        \\
%		topic 36:   &        & topic 36:   &        & topic 36:   &        \\
%		district    & 0.1833 & district    & 0.0279 & district    & 0.0152 \\
%		province    & 0.1726 & school      & 0.0210 & county      & 0.0134 \\
%		county      & 0.0818 & university  & 0.0200 & school      & 0.0096 \\
%		rural       & 0.0813 & province    & 0.0172 & village     & 0.0095 \\
%		village     & 0.0720 & county      & 0.0113 & population  & 0.0089 \\
%		population  & 0.0611 & village     & 0.0107 & province    & 0.0078 \\
%		central     & 0.0573 & population  & 0.0103 & university  & 0.0076 \\
%		families    & 0.0364 & college     & 0.0089 & south       & 0.0068 \\
%		known       & 0.0201 & rural       & 0.0077 & located     & 0.0054 \\
%		formation   & 0.0144 & education   & 0.0074 & central     & 0.0051 \\
		&        &             &        &             &        \\
		topic 37:   &        & topic 37:   &        & topic 37:   &        \\
		born        & 0.1976 & born        & 0.0139 & born        & 0.0007 \\
		american    & 0.0650 & people      & 0.0096 & american    & 0.0006 \\
		people      & 0.0572 & notable     & 0.0092 & street      & 0.0006 \\
		summer      & 0.0484 & american    & 0.0075 & charles     & 0.0004 \\
		notable     & 0.0447 & name        & 0.0031 & said        & 0.0004 \\
		canadian    & 0.0200 & mountain    & 0.0026 & events      & 0.0004 \\
		event       & 0.0170 & japanese    & 0.0025 & people      & 0.0003 \\
		writer      & 0.0141 & fort        & 0.0025 & station     & 0.0003 \\
		dutch       & 0.0131 & character   & 0.0019 & written     & 0.0003 \\
		actor       & 0.0121 & actor       & 0.0014 & point       & 0.0003 \\
		\bottomrule
	\end{tabular}
\end{table}

\section{Conclusion}
We have developed a practical privacy-preserving topic modeling algorithm which outputs accurate and privatized expected sufficient statistics and expected natural parameters. Our approach uses the moments accountant analysis combined with the privacy amplification effect due to subsampling of data, which significantly decrease the amount of additive noise for the same expected privacy guarantee compared to the standard analysis.

\newpage
%-------------------------------------------------------------------------- 
\small
\bibliographystyle{unsrt}
\bibliography{DPEPrefs}

\end{document}